\pdfoutput=1 
\documentclass[sigconf]{acmart}

\AtBeginDocument{%
  }

\setcopyright{acmlicensed}
\copyrightyear{2025}
\acmYear{2025}
\acmDOI{XXXXXXX.XXXXXXX}
\acmConference[Conference acronym 'XX]{ACM Workshop on Informaion Hiding and Multimedia Security}{June 18--20,
  2025}{San Jose, CA USA}
\acmISBN{978-1-4503-XXXX-X/2018/06}

\usepackage{graphicx}
\usepackage{multirow}
\usepackage{hyperref}

\renewcommand{\abstractname}{\MakeUppercase{Abstract}}
\begin{document}

\title{Can GPT tell us why these images are synthesized? Empowering Multimodal Large Language Models for Forensics}

\author{Yiran He}
\affiliation{%
  \institution{Institute of Information Engineering, Chinese Academy of Sciences}
  \city{Beijing}
  \country{China}
}
\email{heyiran@iie.ac.cn}

\author{Yun Cao}
\affiliation{%
  \institution{Institute of Information Engineering, Chinese Academy of Science}
  \city{Beijing}
  \country{China}
  }
\email{caoyun@iie.ac.cn}

\author{Bowen Yang}
\affiliation{%
  \institution{Institute of Information Engineering, Chinese Academy of Science}
  \city{Beijing}
  \country{China}
  }
\email{yangbowen@iie.ac.cn}

\author{Zeyu Zhang}
\affiliation{%
  \institution{Institute of Information Engineering, Chinese Academy of Science}
  \city{Beijing}
  \country{China}
  }
\email{zhangzeyu@iie.ac.cn}

\renewcommand{\shortauthors}{Trovato et al.}

\begin{abstract}
The rapid development of generative AI facilitates content creation and makes image manipulation easier and more difficult to detect. While multimodal Large Language Models (LLMs) have encoded rich world knowledge, they are not inherently tailored for combating AI-generated Content (AIGC) and struggle to comprehend local forgery details. In this work, we investigate the application of multimodal LLMs in forgery detection. We propose a framework capable of evaluating image authenticity, localizing tampered regions, providing evidence, and tracing generation methods based on semantic tampering clues. Our method demonstrates that the potential of LLMs in forgery analysis can be effectively unlocked through meticulous prompt engineering and the application of few-shot learning techniques. We conduct qualitative and quantitative experiments and show that GPT4V can achieve an accuracy of 92.1\% in Autosplice and 86.3\% in LaMa, which is competitive with state-of-the-art AIGC detection methods. We further discuss the limitations of multimodal LLMs in such tasks and propose potential improvements.
\end{abstract}




\maketitle

\begin{figure}[htbp]
  \includegraphics[width=\linewidth]{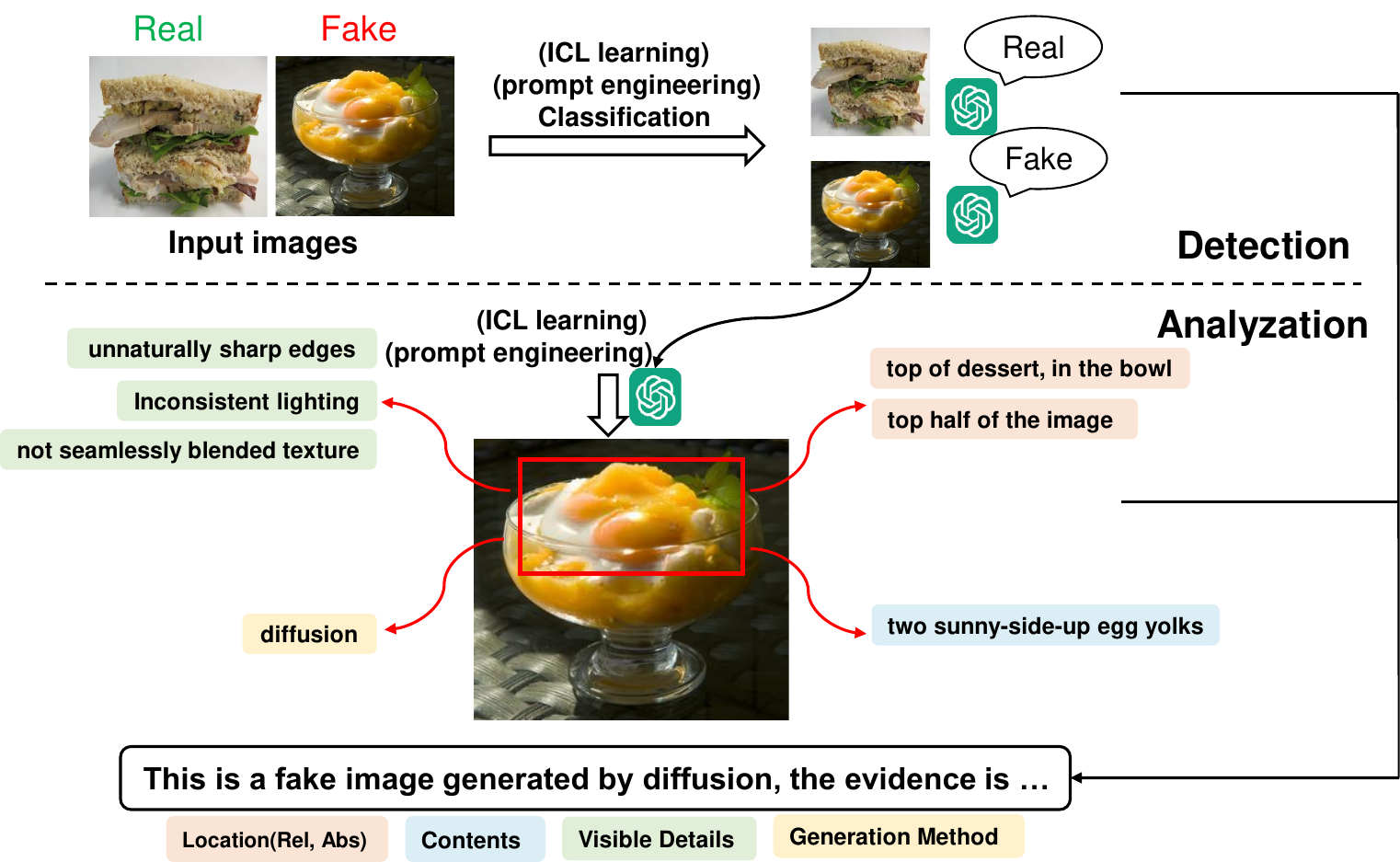}
  \caption{The overall process of leveraging multimodal LLMs to analyze synthesized images. First, we treat it as a fake image classification task. Then, we stimulate LLMs' forensic analyzing ability by prompt engineering and ICL learning. LLMs generate the final report from four perspectives: Location, Contents, Visible Details, and Generation Method.}
  \Description{The overall workflow of the multimodal LLM method.}
  \label{fig:general}
\end{figure}

\section{INTRODUCTION}
Generative Artificial Intelligence (GenAI) has rapidly developed in recent years, enabling the creation of highly realistic synthetic content involving images, audio, and videos from text prompts. While most AI Generated Content (AIGC) benefits humans in fields such as the movie and advertising industry, its misuse has created deleterious content, commonly known as DeepFakes. These AI-generated media, which can convincingly mimic real individuals in both appearance and voice, have raised significant societal concerns. For instance, DeepFake videos have been employed to spread false information from seemingly trusted sources, leading to public confusion and erosion of trust. DeepFakes have been utilized to disseminate misinformation, manipulate public opinion, and infringe upon personal privacy. 

To combat this growing threat, current DeepFake detection methods primarily rely on small-scale machine learning models, specifically Convolutional Neural Networks (CNNs) and optical flow analysis \cite{frank2020leveraging, liu2020global, wang2020cnn}. These approaches focus on identifying artifacts or inconsistencies in manipulated media, such as unnatural facial movements or irregular textures. While these methods have achieved moderate success, they often struggle with generalization across diverse datasets and fail to leverage the contextual understanding that larger, more sophisticated models could provide. This limitation highlights the need for more advanced detection frameworks capable of handling the increasing complexity and realism of DeepFake content.

Meanwhile, Large Language Models (LLMs), such as GPT \cite{gpt} and its successors \cite{llama,deepseek}, have demonstrated remarkable capabilities in natural language processing and content generation. These models, trained on vast datasets, excel at understanding context, semantics, and complex patterns. Recently, the integration of multimodal capabilities into LLMs has expanded their utility beyond text, enabling them to process and analyze images, audio, and video. Multimodal LLMs, such as those combining vision and language, have shown promise in tasks like image captioning, visual question answering, and even forensic analysis. Despite the potential of multimodal LLMs, their application in AIGC content detection is rather limited. Current applications predominantly focus on isolated forgery detection tasks \cite{forgerygpt, fakeshield} or simple question-and-answer interactions ~\cite{jia2024chatgptdetectdeepfakesstudy, liu2024fkaowladvancingmultimodalfake}. These studies exhibit two primary limitations. First, previous efforts in utilizing large models for forgery detection have primarily centered on using certain outputs from the models or features from specific layers as intermediate steps to accomplish particular tasks. This approach necessitates that users identify different types of tampering in advance and employ specific detection methods accordingly, thereby significantly diminishing the practical utility of these models. Second, the utilization of the outputs from large models remains insufficient, concentrating largely on multiple-choice questions, cloze tests, or straightforward Q\&A formats. Such simplistic approaches fail to leverage the semantic information present in the model’s responses and do not capitalize on the models' ability to generate highly interpretable answers.

In this work, we aim to bridge this research gap by stimulating multimodal LLMs' ability in the context of synthesized content detection. While traditional DeepFake detection methods tend to utilize intrinsic features of the image (pixel inconsistencies, frequency domain analysis, and so on.) to identify forged images, LLMs, which are trained on massive corpora, are more inclined to discern images from a semantic perspective, thus resembling human-like interpretation more closely. We structure our approach into two progressive stages as in Figure~\ref{fig:general}, mirroring the cognitive steps a human analyst might take when examining suspicious images. In the first stage, the model assesses whether an image is real or fake based on visual input and a simple prompt. In the second stage, if the image is deemed fake, the model identifies potential reasons for this determination, such as inconsistencies in lighting, texture, or semantic content, and attempts to localize the manipulated regions. In the meantime, the model categorizes the type of forgery and identifies the underlying generation method, GAN or Diffusion in particular. We utilize two strategies to enhance the forgery analyzing ability of LLMs: prompt engineering and In Context Learning(ICL) technology. Our further experiments show that with proper prompt and few-shot learning, LLMs can accomplish these tasks at the same time and show competitive performance with SOTA methods.

By systematically evaluating the capabilities of multimodal LLMs in these tasks, we aim to demonstrate their potential as powerful tools for DeepFake forensics. Our contributions are summarized as follows:
\begin{itemize}
\item Multimodal LLMs can leverage their semantic comprehension to distinguish between authentic and AI-generated images, which comes from their world knowledge gained during pre-training. Unlike traditional machine learning detection methods, LLMs can provide human-interpretable explanations for their decisions, enhancing transparency and trust in the detection process.
\item We carefully crafted our prompts based on five basic principles, and utilized a two-shot ICL strategy in detection and analysis tasks to inspire LLMs' forgery-analyzing ability, which proves effective in further experiments. By correct stimulation methods, multimodal LLMs exhibit the ability to identify and describe manipulated regions within images and trace the methods used for forgery. 
\item Compared with other llm-based forgery analysis methods, our approach can fully leverage the multi-task processing capabilities of LLMs, integrating evidence to provide highly interpretable reports for authentication forgery detection. Our approach achieves an Area Under the Curve (AUC) of 92.1\% in identifying synthesized images and 94.9\% in generation method tracing for the Diffusion-based method. 
\end{itemize}

We hope that this work will advance the understanding of LLMs in synthetic media analysis and pave the way for their broader adoption in combating the pervasive threat of DeepFakes. The remainder of the paper is organized as follows. Section 2 provides an overview of the relevant literature on DeepFake detection and multimodal LLMs. Section 3 presents the methodology of our study. Comprehensive evaluation results and analysis are given in Section 4, and Section 5 concludes the article.
 
\section{RELATED WORKS}

\subsection{Synthesized Image Detection}

Various methods have been developed to distinguish real from synthetic images, primarily by training deep neural networks for binary classification \cite{wang2020cnn, zhao2021multi, gragnaniello2021gan, asnani2023reverse, guo2022robust}. These methods fall into three categories based on feature extraction. Spatial Feature Learning focuses on extracting spatial features from RGB inputs \cite{wang2020cnn, chai2020makes, marra2018detection, schwarcz2021finding, yu2019attributing}, with some approaches relying only on global features \cite{wang2020cnn, marra2018detection, yu2019attributing}, while others emphasize low-level features and local patches for improved detection \cite{zhao2021multi, gragnaniello2021gan, ju2022fusing, chai2020makes, zhang2021thinking,  schwarcz2021finding, yu2022patch, mandelli2022detecting}. Zhao et al. \cite{zhao2021multi} highlight that forgery artifacts persist in high-frequency components, prompting the use of multi-attentional frameworks. Frequency Feature Learning utilizes frequency-domain analysis to detect artifacts from generative models \cite{gragnaniello2021gan, frank2020leveraging, zhang2019detecting, durall2020watch, corvi2023detection, corvi2023intriguing, qian2020thinking, durall2019unmasking}, employing features like spectrum magnitude and 2D-FFT \cite{frank2020leveraging, zhang2019detecting, qian2020thinking, durall2019unmasking}. However, these methods often rely on fixed filters, limiting adaptability to unseen models and post-processing effects. Feature Fusion integrates multiple complementary features for robust AI-synthesized image detection \cite{chen2021robust, zhang2021thinking, mandelli2022detecting, qian2020thinking}. Techniques include dual-color fusion (RGB and YCbCr) \cite{chen2021robust} and frequency-spatial feature fusion \cite{qian2020thinking}. Unlike previous methods, our approach introduces LLM-based detection, improving generalization against advanced generative models and diverse forgery types.

\subsection{Multimodal Large Language Models}

LLMs, such as GPT-3 \cite{gpt3}, LLaMA \cite{llama}, and DeepSeek \cite{deepseek}, have demonstrated remarkable performance across a wide range of natural language processing tasks. More recently, researchers have been exploring ways to extend LLMs' capabilities to multimodal domains, enabling them to perceive and reason about visual signals. Pioneering efforts, such as LLaVA \cite{llava} and Mini-GPT4 \cite{minigpt4}, focus on aligning image and text features, followed by visual instruction tuning. This process involves additional training of pre-trained models using curated instruction-formatted datasets to improve their generalization to unseen tasks. Similarly, PandaGPT \cite{pandagpt} introduces a simple linear projection layer to bridge ImageBind \cite{imagebind} and Vicuna \cite{vicuna}, allowing for multimodal inputs. The success of multimodal LLMs has catalyzed research in various specialized domains, including medical applications \cite{li2023llava}, video understanding \cite{jin2024chat,liu2023devan}, and image editing \cite{fu2023guiding}. 

Given the increasing sophistication of generative models, multimodal LLMs have also emerged as a promising tool for forensic applications, particularly in detecting and analyzing synthetic images. FKA-Owl \cite{liu2024fkaowladvancingmultimodalfake} enhances LLMs for multimodal fake news detection by incorporating forgery-specific knowledge, such as semantic correlation and visual artifact analysis. However, its classification approach is largely confined to binary decision-making and cannot provide detailed linguistic explanations. On the other hand, ForgeryGPT \cite{forgerygpt} integrates a Mask-Aware Forgery Extractor, which improves pixel-level analysis of manipulated images and facilitates interpretable reasoning through multi-turn dialogues. Despite these advancements, the generalizability of ForgeryGPT across diverse manipulation techniques remains an open challenge. In this work, we build upon these advancements by leveraging the world knowledge embedded in multimodal LLMs to enhance the forensic analysis of open-world synthetic images. Our approach aims to provide not only accurate detection but also comprehensive textual explanations, bridging the gap between forensic image analysis and interpretable AI-driven insights.

\section{METHODOLOGY}

\subsection{Architecture Overview}
\begin{figure}[htbp]
  \centering
  \includegraphics[width=\linewidth]{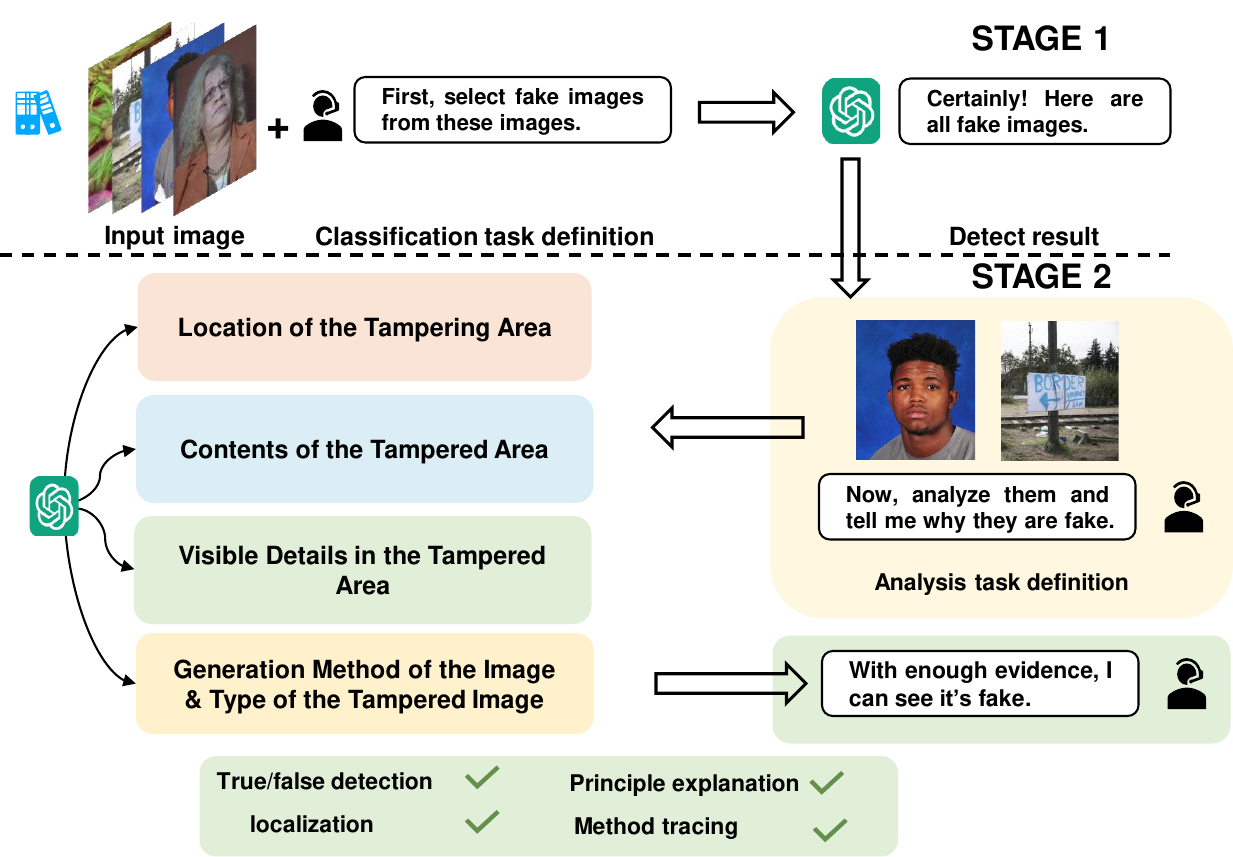}
  \caption{The overall framework of our proposed multimodal LLM forensic analysis framework. By leveraging a two-stage workflow, we can use LLMs once and for all in different types of tasks: (1) localizing the manipulated regions, (2) describing the forged objects, (3) providing reasons for the forgery judgment, and (4) tracing the forgery method.  }
  \Description{The overall workflow of the multimodal LLM method.}
  \label{fig:overall}
\end{figure}

Our goals involve two issues: 1): Utilizing the textual understanding ability and world prior knowledge of the LLMs to analyze and judge the authenticity of tampered images; and 2): Adopting the analysis and interpretation ability of LLMs to assist people in pinpointing the tampered areas. To solve these two tasks, an intuitive approach is to prompt or fine-tune a large multimodal model to simultaneously output detection and analysis. However, we find that joint training of multiple tasks will increase the difficulty of network optimization and interfere with each other. Considering that detection focuses more on language understanding while analyzing requires more accumulation of visual prior information and language generation, the proposed method contains two key decoupled parts, as illustrated in Figure~\ref{fig:overall}. In the first stage, a binary classification task is performed. For an unknown image, the multimodal LLM leverages its pre-trained knowledge and few-shot examples to determine whether the image is real or fake. In the second stage, once images are identified as fake, the LLM is tasked with four further subtasks: (1) localizing the manipulated regions, (2) describing the forged objects, (3) providing reasons for the forgery judgment, and (4) tracing the forgery method.

We decide to use the two-stage strategy for several reasons. First, this structure is designed to align with human cognitive processes. Since LLMs are trained on human language corpora rather than specific datasets of traditional machine learning models, they tend to interpret images semantically rather than at the signal level \cite{wu2025interpreting}. This semantic understanding mirrors human behavior when analyzing potentially forged images: humans first make a rough judgment about the image's authenticity and then scrutinize details to support their initial assessment. Besides, there are many successful cases of a two-stage approach in forensic detection, such as FakeShield \cite{fakeshield}, ForgeryGPT \cite{forgerygpt} and ProFact-NET \cite{profactnet}, which confirms the feasibility of our approach. Additionally, our experiments reveal that longer prompts (prompts with more tokens) tend to increase the likelihood of the LLM classifying an image as fake, a phenomenon consistent with findings on model hallucination in other studies. On the other hand, the study of Shan Jia et al. \cite{jia2024chatgptdetectdeepfakesstudy} has shown that shorter prompts proved useful in the classification of real and fake images. Based on these reasons, we argue that this "judge first, analyze next" approach better harnesses the potential of multimodal LLMs without damaging their performance. Our experiments further show that this approach demonstrates satisfactory performance across real, GAN-generated, and Diffusion-generated images.

\begin{figure}[htbp]
  \centering
  \includegraphics[width=\linewidth]{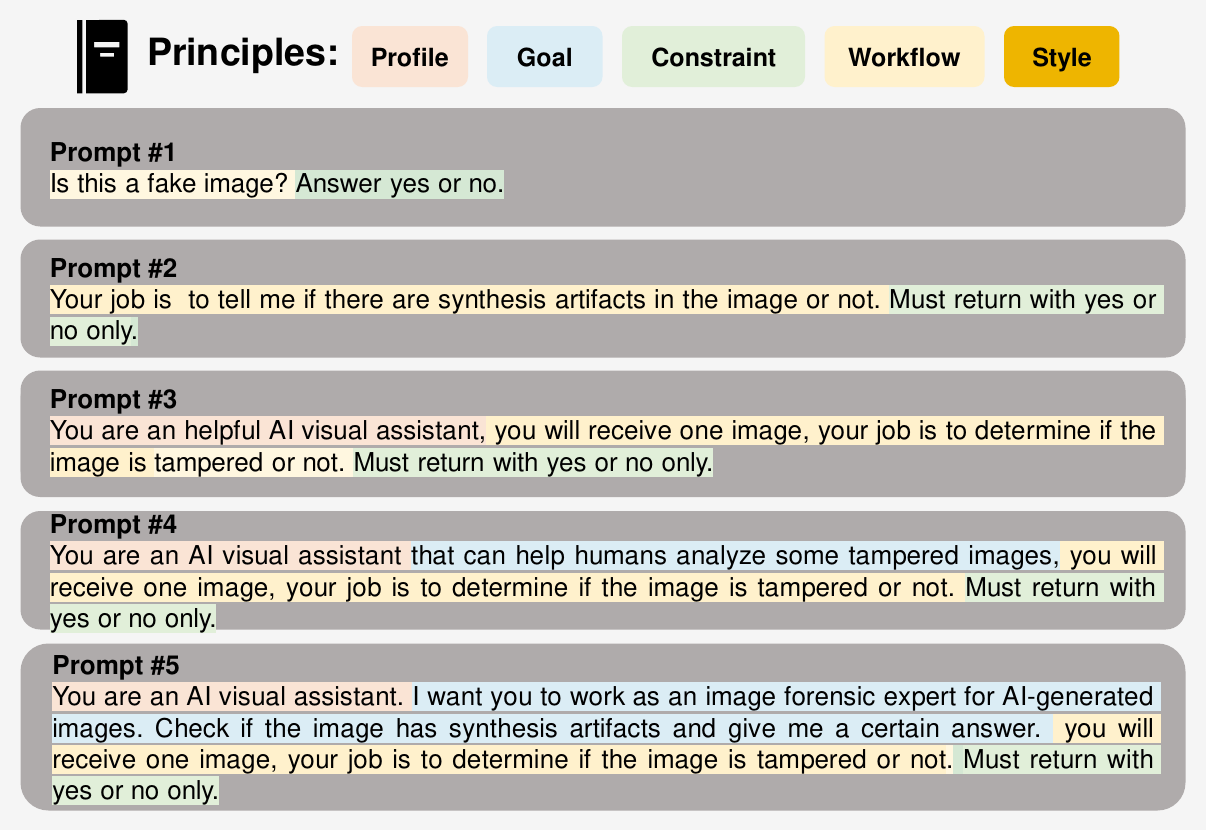}
  \caption{A list of prompts for GPT4V in detecting 1,000 faces from the Autosplice dataset. At the top, we show that the design of all five prompts is based on five basic principles: Profile, Goal, Constraint, Workflow, and Style. From top to bottom, prompts are getting longer and longer, adding more and more principles. We use Prompt \#4 in practice.}
  \Description{Ablation Study of Prompts.}
  \label{fig:prompt_ablation}
\end{figure}

\subsection{Text Prompts}
Text prompts play a crucial role in guiding multimodal LLMs to detect DeepFake images. These prompts consist of instructions and requests designed to leverage the semantic knowledge embedded in the LLMs. Prior research \cite{jia2024chatgptdetectdeepfakesstudy,liu2024fkaowladvancingmultimodalfake} has shown that simplistic prompts are often ineffective. In many cases, LLMs either provide inaccurate responses or refuse to answer due to a lack of contextual information or for safety considerations, especially when dealing with human faces. For instance, when prompted with, "Tell me the probability of this image being AI-generated. Answer a probability score between 0 and 100", GPT-4o exhibits a rejection rate of 80\%. Generally, prompts with richer contextual information tend to reduce the rejection rate of LLMs. However, overly detailed prompts can lead to lower accuracy, as they may cause the model to overemphasize specific cues mentioned in the prompt while ignoring other potentially relevant clues not explicitly stated. This phenomenon aligns with findings on model hallucination, where LLMs generate plausible but incorrect responses based on biased input.

\begin{figure}[htbp]
  \centering
  \includegraphics[width=\linewidth]{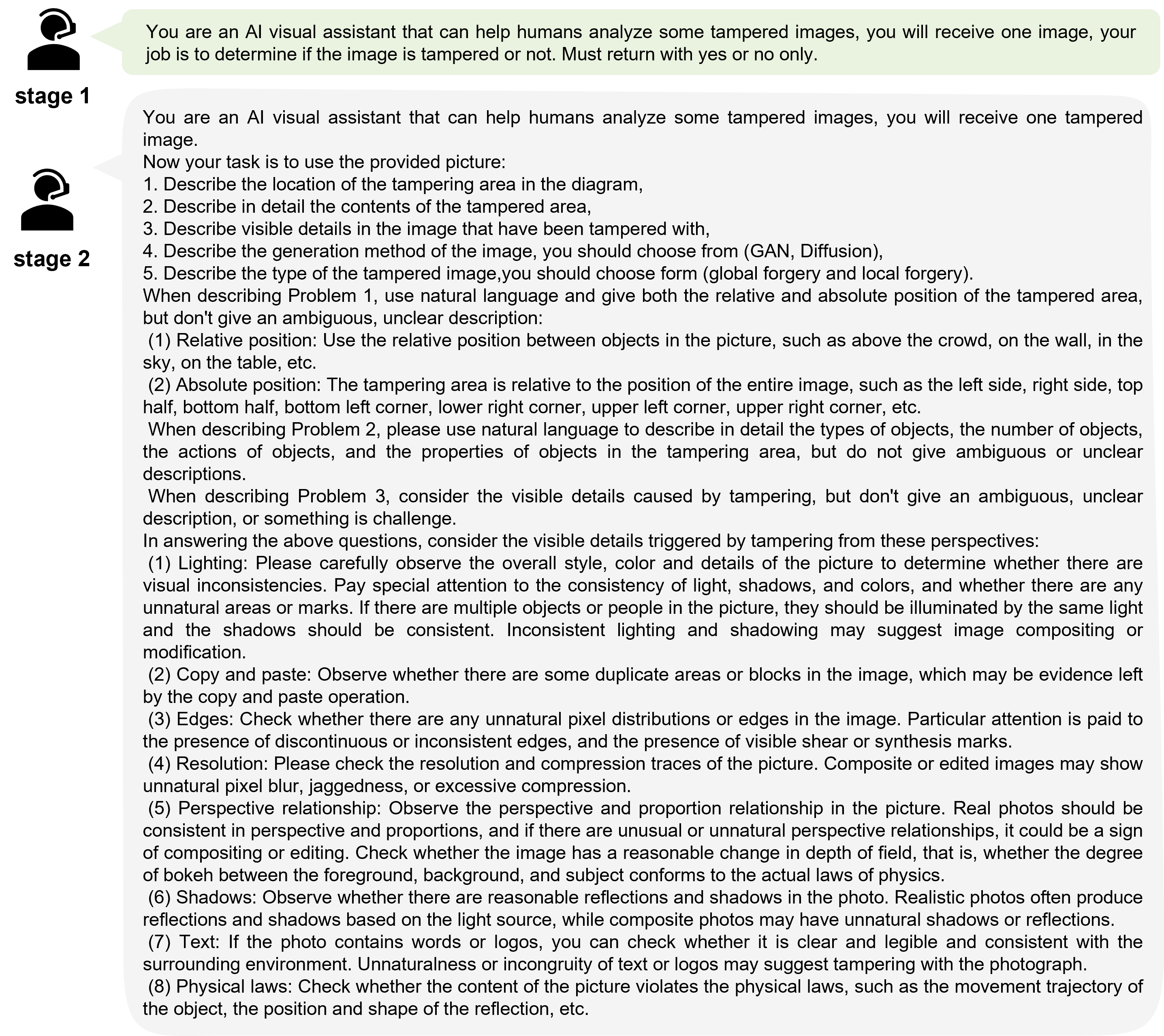}
  \caption{Prompts for GPT-4o when analyzing DeepFakes. In Stage 1, we use a simple prompt to let the llm answer a two-class question; in Stage 2, once recognizing an image as DeepFake, it must analyze the fake image from 4 perspectives: localization, description, reasoning, and tracing. In this process, we provide GPT with as many perspectives for consideration as possible. We also use two examples in the user prompt to inspire the ICL ability of the LLM, which is not shown here. }
  \Description{Prompts for GPT-4o when analyzing DeepFakes.}
  \label{fig:prompt}
\end{figure}

To address these challenges, we carefully designed our prompts to strike a balance between providing sufficient context and avoiding excessive details. As illustrated in Figure~\ref{fig:prompt_ablation}, our prompt design is based on five principles: Profile, Goal, Constraint, Workflow, and Style, which come from the inspiration of the design of LangGPT \cite{wang2024langgpt}. In the process of designing the prompt, we also refer to the OpenAI official documentation regarding prompt design\footnote{\url{https://platform.openai.com/docs/guides/prompt-engineering}}. Our final prompt design is demonstrated in Figure~\ref{fig:prompt}. In Stage 1, we use simple binary prompts to ask for straightforward Yes/No answers. In Stage 2, we go beyond simple binary answers: we ask the LLM to first localize the area of synthesis and then make a short description of it. In the meantime, we request it to describe visible details in the image that have been tampered with and further trace the generation method and type of the tampered image. This additional request can lead the LLM to be more guided, resulting in the lowest rejection rate. Based on the evidence above, LLMs can finally judge the authenticity of the image. Using these prompts, we guide the LLMs through a structured analysis process, mimicking human forensic workflows while minimizing the risk of refusal or hallucination. More reasons why we chose these prompts are illustrated in Ablation Studies. Additionally, to standardize the outputs of the LLMs and better harness their potential, we employ ICL techniques to provide examples for LLMs. Subsequent experiments show that the use of examples increases the model's accuracy by approximately 12\% and significantly reduces the rejection rate.

\subsection{Stage 1: Forgery Detection}
\begin{figure}[htbp]
  \centering
  \includegraphics[width=\linewidth]{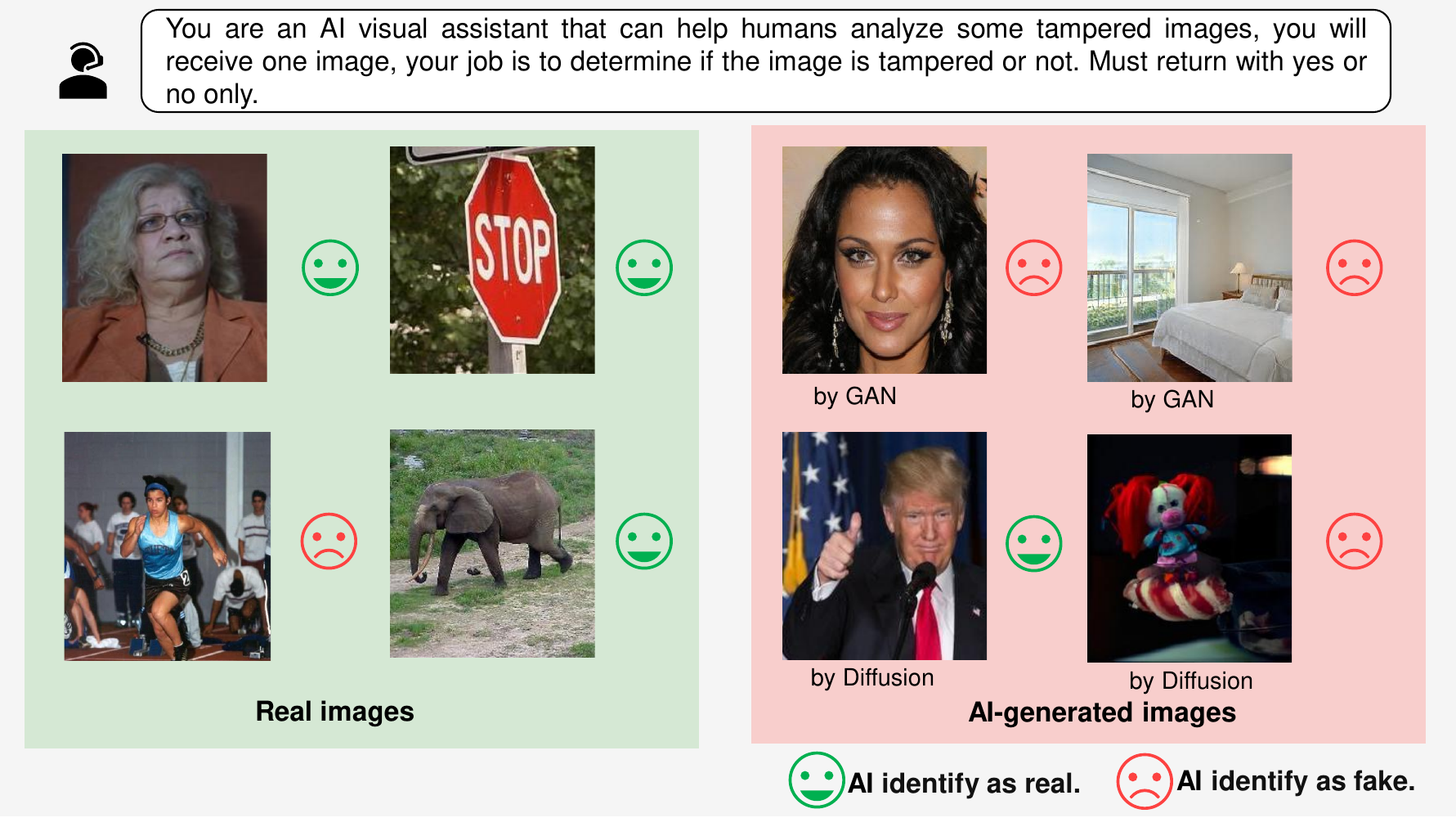}
  \caption{Examples of GPT-4o for DeepFake classification in Stage 1, containing both objects and human faces. Left: Results for real images from the Caltech-101 \cite{caltech101} dataset and the Caltech-WebFaces \cite{fink_perona_2022} dataset. Right: Results for AI-generated images from Stable Diffusion \cite{stablediffusion} and StyleGAN \cite{stylegan} dataset. The responses for real faces are labeled in green, while those for AI-generated faces are labeled in pink. Both success (with a happy icon) and failure (with an unhappy icon) are shown.}
  \Description{Examples of GPT-4o when analyzing DeepFakes in Stage 1.}
  \label{fig:s1example}
\end{figure}

In real-world scenarios, images can be manipulated or subjected to various forms of attacks, including splicing, object removal, DeepFake generation, and AI-generated content (AIGC) techniques. However, these tampered images exhibit diverse distribution characteristics and domain-specific variations, posing significant challenges for a single detection method to comprehensively capture all their features. At the same time, LLMs are trained on extensive human corpora and possess advanced world knowledge and semantic understanding. By leveraging these capabilities, LLMs can assist in forensic analysis, mitigating the limitations of conventional detection methods and enhancing the robustness of image authenticity verification. In Figure~\ref{fig:s1example}, we present several examples of binary classification results using GPT-4V in Stage 1. The left column represents real images, while the right column showcases DeepFake images generated by GAN or Diffusion methods. Successful cases are marked with a happy icon, and failures are indicated with an unhappy icon. In Stage 1, the assistant only answers yes/no
results without supporting evidence. In the user prompt, we use a real example and a fake example to unlock the potential of large models and reduce the rejection rate.

\subsection{Stage 2: Forgery Analysis}
\begin{figure}[htbp]
  \centering
  \includegraphics[width=\linewidth]{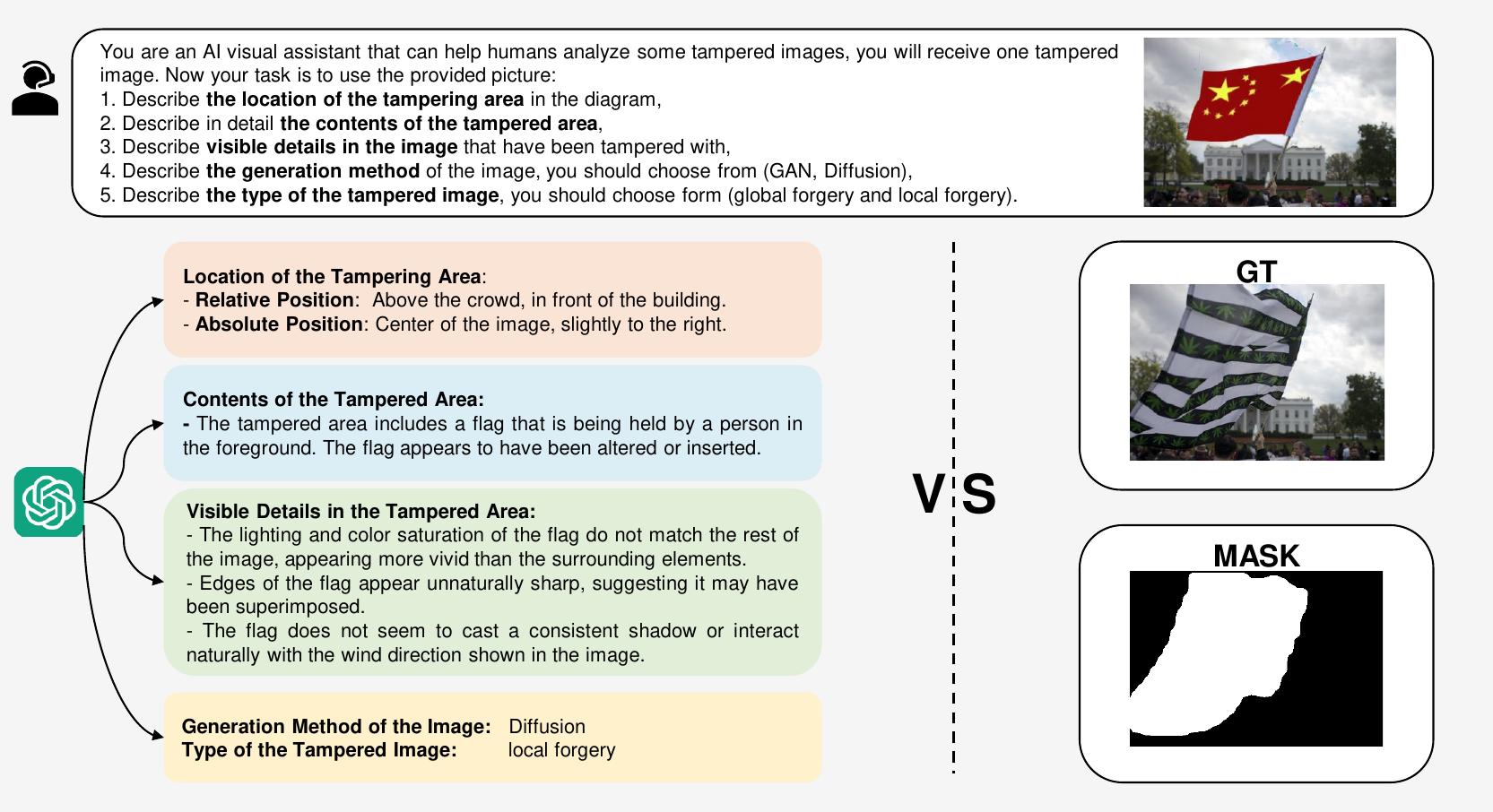}
  \caption{An example of GPT-4o for DeepFake analysis in Stage 2. Left above: Human prompts for DeepFake analysis. We let LLM do all tasks in a single round. Right above: a DeepFake image. This one comes from the Autosplice dataset, which consists of local forgery images generated by the Diffusion method. Left bottom:  Answers from the LLM. We use different colors as a background for different tasks. Right bottom: original image (GT) and mask of the DeepFake image, which will be used in the evaluation of the localization task later.}
  \Description{Examples of GPT-4o when analyzing DeepFakes in Stage 2.}
  \label{fig:s2example}
\end{figure}

Although LLMs demonstrate the capability to detect forged data in the first stage, they cannot still precisely describe the manipulated regions. To address this limitation, in the second stage, we enhance the model’s capacity to provide valid arguments supporting the classification of an image as falsified. Existing forensic methods based on LLMs \cite{jia2024chatgptdetectdeepfakesstudy,liu2024fkaowladvancingmultimodalfake} fail to generate detailed descriptive information and struggle to describe the semantic content of images. we leverage the semantic understanding capabilities of multimodal LLMs in Stage 2 to perform a detailed analysis of DeepFake images. This stage focuses on extracting and analyzing specific features of the forged images, providing both qualitative and quantitative evaluations. The key information we aim to extract from the images includes:
\begin{itemize}
\item Location of the Tampering Area: Identifying where the image has been manipulated.
\item Contents of the Tampered Area: Describing the objects or elements within the manipulated region.
\item Visible Details in the Tampered Area: Highlighting specific visual anomalies or inconsistencies.
\item Generation Method and Type of the Image: Determining the forgery technique used (e.g., GAN or Diffusion) and the type of forgery (e.g., global or local).
\end{itemize}
Figure~\ref{fig:s2example} visually demonstrates an example of Stage 2 analysis for a forged image. At the top of the figure, we show the model's input, including the forged image and the prompt (as described in Figure~\ref{fig:prompt}), along with a few example responses to standardize the LLM's output format. The model's detection results are displayed in the bottom-left corner, with different colors representing outputs for different tasks. For the localization task, the LLM provides both relative and absolute positions of the tampered regions, this output format facilitates subsequent verification using the original image (GT) and the mask image (MASK). For the tampered area description task, the LLM describes the type, quantity, and behavior of the forged objects. As shown in Figure~\ref{fig:s2example}, the model identifies the flag in the foreground as the primary forged object. 
For the visible details task, the LLM provides reasoning for its forgery judgment, such as identifying suspicious attributes like lighting, edges, and shadows of the flag in the image. Finally, for the generation method and type task, the LLM predicts the forgery technique (e.g., GAN or Diffusion) and the type of forgery (e.g., global or local).

\section{EXPERIMENT}

\subsection{Experimental Setup}

\textbf{Dateset:} Our dataset comprises 1,000 real general images sourced from the Caltech-101 \cite{caltech101} dataset and 1,000 real face images from the Caltech-WebFaces \cite{fink_perona_2022} dataset. For forged images, we included 4,000 globally manipulated images generated using Stable Diffusion \cite{stablediffusion} and StyleGAN \cite{stylegan}, as well as 4,000 locally manipulated images from AutoSplice \cite{autosplice} and LaMa \cite{LaMa}. Despite general images, we also collect two forgery face datasets, AutoSplice \cite{autosplice} and HiSD \cite{HiSD}, with 1000 images each. These datasets cover two prominent AI generation methods: GANs and Diffusion models, ensuring a comprehensive evaluation of the LLMs' capabilities.

\noindent \textbf{State-of-the-Art Methods:} To contextualize the performance of multimodal LLMs in DeepFake detection, we select some state-of-the-art methods and utilize them in our forgery detection task. Specifically, we compare our method with three state-of-the-art approaches: FreDect \cite{frank2020leveraging}, GramNet \cite{liu2020global}, and CNNSpot \cite{wang2020cnn}, each addressing DeepFake detection from a different perspective. FreDect utilizes frequency-based analysis to effectively identify DeepFake images by detecting artifacts introduced during the generation process. GramNet employs a deep learning architecture that enhances global texture representations, improving the robustness and generalization of fake face detection across different datasets. CNNSpot demonstrates that CNN-generated images retain distinct artifacts, enabling a classifier trained on a single model (e.g., ProGAN) to generalize well to other architectures. All of these models are trained in the datasets specified by their respective authors and tested in our evaluation datasets.

\begin{table}
  \caption{Comparsion of ACC (\%) in detecting DeepFake general images. "Stable" stands for the Stable Diffusion \cite{stablediffusion} model, and "Style" represents the StyleGAN \cite{stylegan} model.}
  \label{tab:2cls acc}
\resizebox{\linewidth}{!}{
\begin{tabular}{@{}lccccc@{}} \toprule
 \multirow{2}{*}{Method} & \multirow{2}{*}{Real} & \multicolumn{2}{c}{Diffusion} & \multicolumn{2}{c}{GAN}\\ \cmidrule(l){3-6}
 & & Autosplice\cite{autosplice} & Stable\cite{stablediffusion} & LaMa\cite{LaMa} & Style\cite{stylegan} \\ \midrule
 FreDect\cite{frank2020leveraging} & 95.3 & \textbf{20.5} & 9.2 & \textbf{57.2} & \textbf{94.3} \\
 GramNet\cite{liu2020global} & \textbf{100.0} & 5.3 & \textbf{9.2} & 0.1 & 3.9 \\
 CNNSpot\cite{wang2020cnn} & 99.0 & 0.2 & 3.1 & 1.2 & 64.2 \\ \midrule \midrule
 Deepseek & 55.2 & 76.3 & 79.3 & 77.3 & \textbf{84.8} \\
 Llama & 71.3 & 83.5 & \textbf{86.3} & 77.4 & 80.6 \\
 GPT4V & \textbf{87.1} & \textbf{92.1} & 84.3 & \textbf{86.3} & 73.3 \\ \bottomrule
 \end{tabular}
}
\end{table}

\noindent \textbf{Detection Metrics:} To evaluate the performance of multimodal LLMs in the proposed two-stage DeepFake detection framework, we employ a set of robust metrics tailored to the unique characteristics of LLM outputs. For binary classification tasks, such as determining whether an image is real or fake (Stage 1) or identifying the forgery method as GAN or Diffusion (Stage 2), we adopt a probabilistic scoring approach. Specifically, for each text-image prompt, we query the LLM multiple times and compute the average score based on the model's responses (e.g., assigning No = 0 and Yes = 1). This approach offers two key advantages. First, since LLMs generate tokens probabilistically and employ a top-k strategy to select outputs, averaging multiple responses helps assess the diversity and consistency of the model's answers to the same query. Second, using numerical decision scores enables us to extend performance evaluation beyond simple accuracy (ACC) to more comprehensive metrics such as the Receiver Operating Characteristic (ROC) curve and the Area Under the Curve (AUC) score. Unlike ACC, AUC is not affected by class imbalance, providing a more reliable assessment of model performance. Additionally, AUC allows for direct comparison with existing programmatic detection methods, facilitating a broader evaluation of LLM capabilities in forensic tasks.  

\noindent \textbf{Forgery Location Metrics:} For local forgery tasks, we introduce an additional evaluation metric to assess the LLM's ability to accurately localize manipulated regions using another LLM as a judge. With the source image and the corresponding MASK image as ground truth, we evaluate the model's performance in identifying and describing the forged areas, as shown in Figure~\ref{fig:location_example}. The model's output includes both absolute and relative positions of the artifacts, expressed through natural language descriptions. To quantify localization accuracy, The evaluation metrics are divided into four components: Absolute Position Accuracy, Relative Position Accuracy, Readability, and Completeness. The average of these scores is then calculated to determine the model's localization accuracy. This approach allows us to measure how effectively the LLM can pinpoint manipulated regions, providing insights into its spatial reasoning capabilities in forensic tasks.

\begin{figure}[htbp]
    \centering
    \begin{minipage}[t]{0.49\linewidth}
        \centering
        \includegraphics[width=\linewidth]{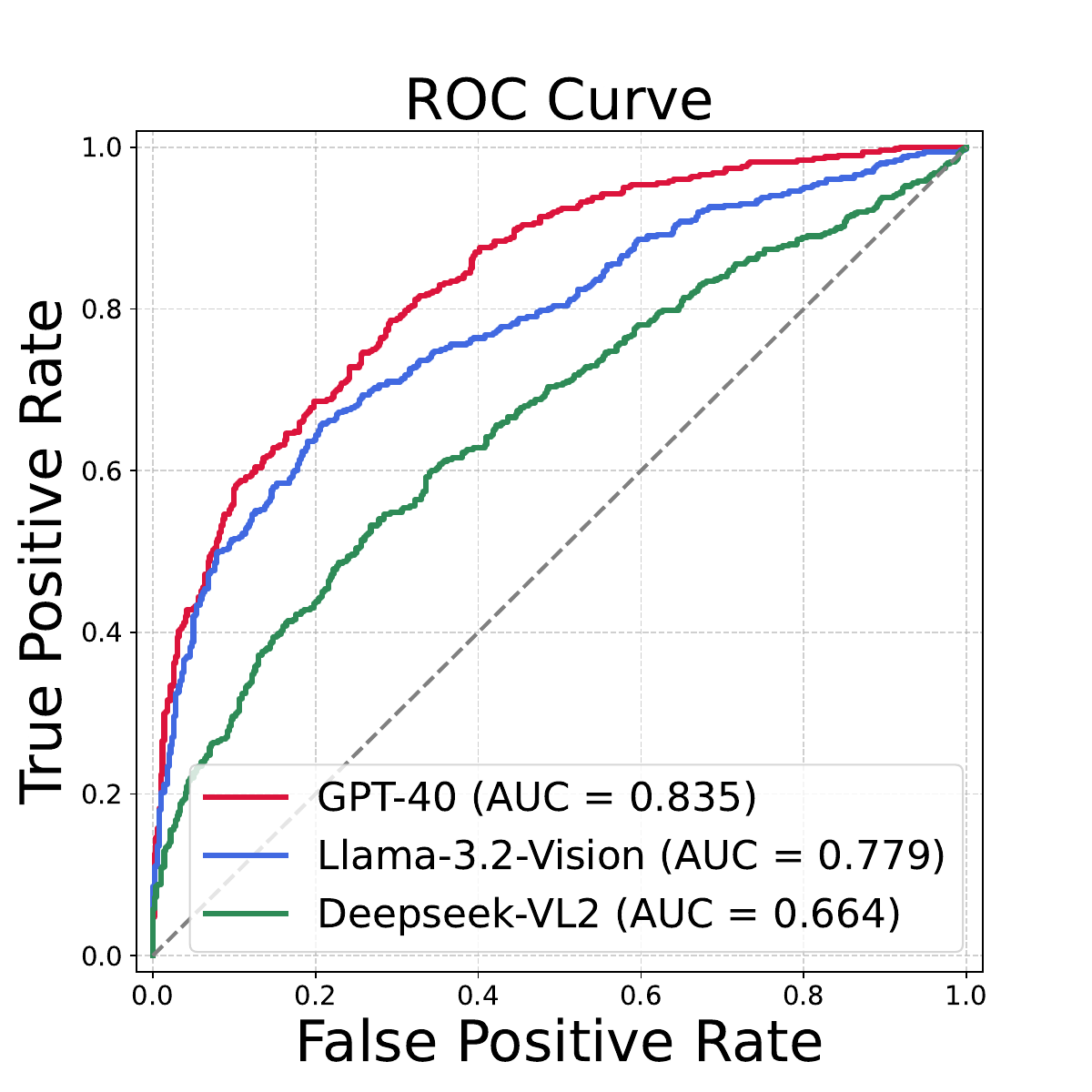}
    \end{minipage}
    \hfill 
    \begin{minipage}[t]{0.49\linewidth}
        \centering
        \includegraphics[width=\linewidth]{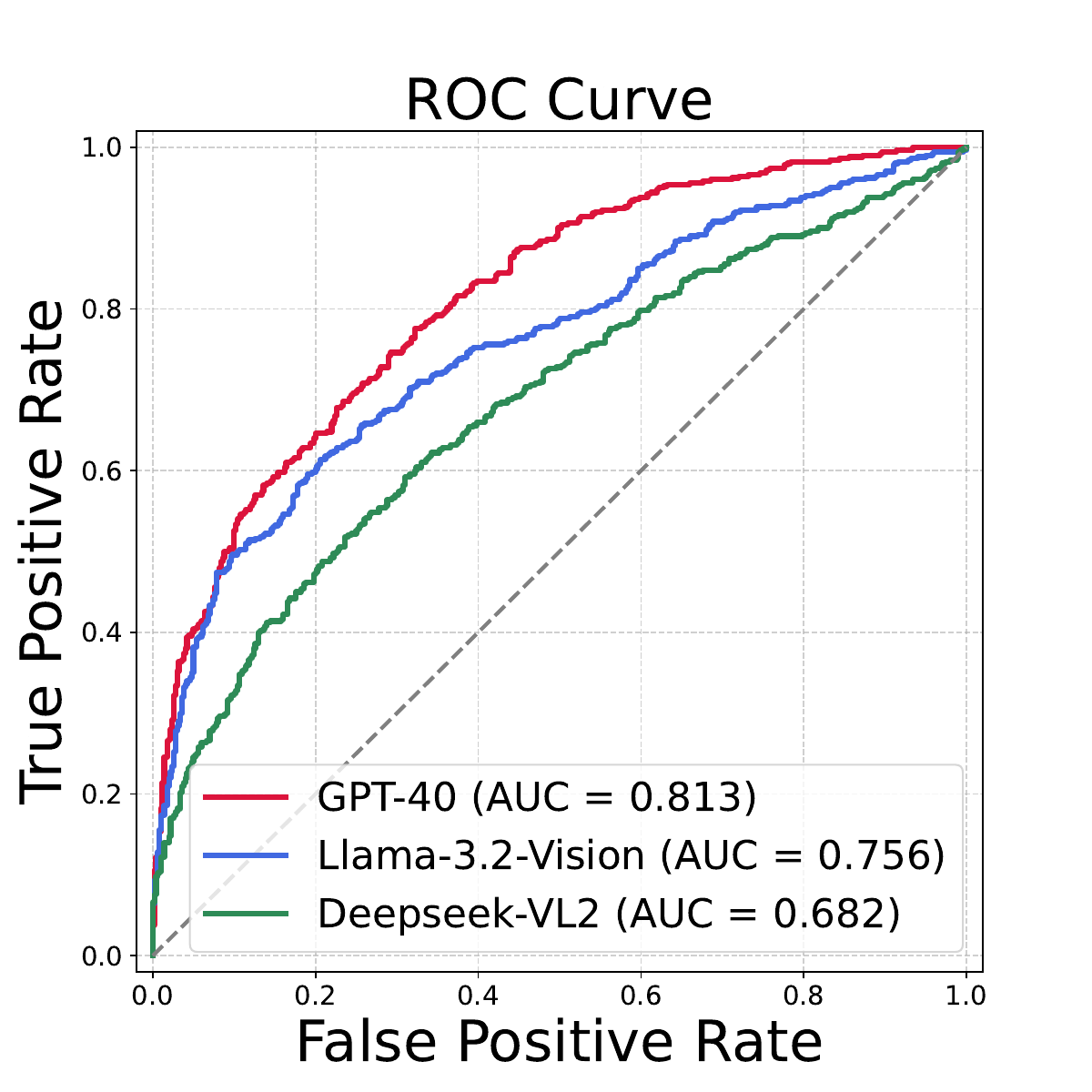}
    \end{minipage}
\caption{ROC curves of three multimodal LLMs (GPT-4O, Llama-3.2-Visiom, Deepseek-VL2) on the DeepFake detection based on averaging the predictions of five rounds of queries, left: on the Diffusion dataset, right: on the GAN dataset.}
\label{fig:ROC}
\end{figure}

\noindent \textbf{Implementation details:} We selected OpenAI's GPT-4 Vision model (gpt-4o-2024-08-06) as the primary model for this study. Its API support for Python enables large-scale simulation of conversational contexts, which is crucial for our experimental design. We also incorporated two open-source LLMs for comparative analysis: Llama-3.2-Vision and DeepSeek-VL2. Due to the limitations of hardware resources, we deploy their smaller visions locally, Llama-3.2-11B-Vision and Deepseek-vl2-small(2.8B) in particular. These models were chosen to provide a broader perspective on the capabilities and limitations of different sizes and structures of multimodal LLMs in forensic tasks. For prompt engineering, we use the same prompt as in Figure~\ref{fig:prompt} and two-shot learning technology for all LLMs. Our two-shot examples also stay the same regardless of the change in datasets and models. For GPT-4o, all our evaluations were conducted through API calls. We adhered to the default parameter settings as specified on the official OpenAI API website\footnote{\url{https://platform.openai.com/docs/models\#gpt-4o}}. In contrast, we deployed the llama\footnote{\url{https://www.llama.com/docs/model-cards-and-prompt-formats/llama3_2/\#-llama-3.2-vision-models-(11b/90b)-}} and deepseek\footnote{\url{https://github.com/deepseek-ai/DeepSeek-VL}} models locally and conducted our evaluations following their official guides. The total cost for Openai API calls was approximately \$150, and the study took around 40 days. 


\subsection{Forgery Detection Performance}

The qualitative results demonstrate that the LLMs achieves a reasonable level of accuracy in distinguishing real images from AI-generated ones, and quantitative results further support this observation. Figure~\ref{fig:ROC} illustrates the ROC curves and the AUC scores obtained using our designed prompts on the evaluation dataset. GPT-4V achieves an AUC of 83.5\% for Diffusion-generated images and 81.3\% for GAN-generated images. These results confirm that GPT-4V is not performing random guessing, which would correspond to a diagonal ROC curve with an AUC of 50\%. In comparison, Llama-3.2-Vision shows a slight performance drop, with an AUC of 77.9\% for Diffusion-generated images and 75.6\% for GAN-generated images. DeepSeek-VL2 exhibits a more noticeable decline in performance, with AUC scores approximately 10\% lower than Llama-3.2-Vision. However, its performance still significantly surpasses random guessing, indicating its capability to distinguish real from fake images, albeit with reduced accuracy.

\begin{table}
  \caption{Comparsion of AUC (\%) in detecting DeepFake general images. "Stable" stands for the Stable Diffusion \cite{stablediffusion} model, and "Style" represents the StyleGAN \cite{stylegan} model.}
  \label{tab:2cls auc}
\resizebox{\linewidth}{!}{
\begin{tabular}{@{}lcccc@{}} \toprule
 \multirow{2}{*}{Method} & \multicolumn{2}{c}{Diffusion} & \multicolumn{2}{c}{GAN}\\ \cmidrule(l){2-5}
  & Autosplice\cite{autosplice} & Stable\cite{stablediffusion} & LaMa\cite{LaMa} & Style\cite{stylegan} \\ \midrule
 FreDect\cite{frank2020leveraging} & \textbf{56.8} & 53.2 & \textbf{73.5} & \textbf{94.2} \\
 GramNet\cite{liu2020global} & 52.6 & \textbf{54.3} & 50.1 & 52.3 \\
 CNNSpot\cite{wang2020cnn} & 50.1 & 51 & 49.6 & 81.6 \\ \midrule \midrule
 Deepseek & 65.1 & 67.6 & 66.3 & 70.5 \\
 Llama & 79.4 & 77.8 & 73.3 & \textbf{78.9} \\
 GPT4V & \textbf{84.2} & \textbf{85.6} & \textbf{83.0} & 77.4 \\ \bottomrule
 \end{tabular}
 }
\end{table}

In Table~\ref{tab:2cls acc}, we present the accuracy (ACC) comparison, and in Table~\ref{tab:2cls auc}, we compare the AUC scores. To further validate the differences between LLM-based detection and traditional methods, we analyze the results separately for Diffusion-generated and GAN-generated datasets. Specifically, we include two GAN-based datasets: LaMa (local generation) and StyleGAN (global generation), and two Diffusion-based datasets: AutoSplice (local generation) and Stable Diffusion (global generation). As shown in the tables, the performance of the three multimodal LLMs, GPT-4V, Llama-3.2-Vision, and DeepSeek-VL2 follows a descending order, with GPT-4V achieving the highest scores. This trend can be attributed to the significant difference in model size: GPT-4V (potentially 1T parameters) vastly outperforms Llama-3.2-Vision (11B ) and DeepSeek-VL2 (2.8B), suggesting a positive correlation between model size and DeepFake detection capability.

Besides, we note that traditional DeepFake detection methods exhibit strong performance on real datasets and some specific forgery datasets but struggle with others. In contrast, all LLM-based methods demonstrate more balanced accuracy across datasets, except for DeepSeek-VL2, which achieves only 55.2\% ACC on real datasets. This discrepancy highlights a fundamental difference between the two approaches. Traditional methods rely on capturing signal-level discrepancies between real and AI-generated images during training. When encountering unseen data, these methods often fail because the image characteristics differ from the training set, rendering the pre-trained classifiers ineffective. In contrast, LLMs base their decisions on semantic-level anomalies, as evidenced by the natural language explanations provided in Stage 2. Despite not being explicitly trained for DeepFake detection, LLMs leverage their internal world knowledge to perform this task effectively. However, we observe that LLMs tend to make more errors on real images, particularly DeepSeek-VL2, which shows an ACC gap of over 40\% than traditional DeepFake detection methods. This may be due to the models misinterpreting "unusual" features (e.g., motion blur or camera focus issues, as seen in the bottom-left image of Figure~\ref{fig:s1example}) as signs of forgery. This suggests that the semantic anomalies identified by LLMs may sometimes conflict with real-world scenarios, a limitation that could be addressed through refining the model.

Additionally, as shown in Table~\ref{tab:2cls auc}, when the generation method shifts from global to local forgery, traditional DeepFake detection methods exhibit significant performance fluctuations (e.g., FreDect's AUC drops from 94.2\% to 73.5\%). In contrast, LLMs are less affected by this change. This is because local forgery retains many regions of the original image, making signal-level differences less pronounced and confusing traditional methods. LLMs, however, rely on semantic inconsistencies, which are still present in locally forged images, enabling them to detect manipulations effectively.

\begin{table}
  \caption{Comparsion of ACC (\%) in detecting DeepFake faces. Autosplice\cite{autosplice} is a Diffusion-based dataset, and HiSD\cite{HiSD} is a GAN-based dataset.}
  \label{tab:face acc}
\begin{tabular}{@{}lccc@{}} \toprule
 Method & Real & Autosplice\cite{autosplice} & HiSD\cite{HiSD} \\ \midrule
 FreDect\cite{frank2020leveraging} & 92.2 & 33.4 & \textbf{67.6}  \\
 GramNet\cite{liu2020global} & \textbf{100} & \textbf{43.9} & 0 \\
 CNNSpot\cite{wang2020cnn} & 100 & 1.1 & 13.1 \\ \midrule \midrule
 Deepseek & 36.1 & 68.1 & 70.6 \\
 Llama & 36.3 & 67.2 & 63.3 \\
 GPT4V & \textbf{76.7} & \textbf{79.6} & \textbf{76.2} \\ \bottomrule
 \end{tabular}
\end{table}

To investigate how multimodal LLMs perform on face images compared to general images, we conducted experiments using a real-face dataset (Caltech-101), a Diffusion-based dataset (AutoSplice), and a GAN-based dataset (HiSD). We compared the performance of traditional Deepfake detection methods and LLMs on these face datasets, as summarized in Table~\ref{tab:face acc}. We can see that traditional Deepfake detection methods achieve high accuracy on real-face datasets but show inconsistent performance on forged datasets. This aligns with the findings for general images, where traditional methods excel on specific datasets but struggle with others due to their reliance on signal-level features. Meanwhile, compared to their performance on general images (as shown in Table~\ref{tab:2cls acc}), LLMs exhibit lower accuracy on face datasets. This suggests that face forgery detection is more challenging for LLMs than general image forgery detection. An intuitive reason is that faces are influenced by numerous factors, such as age, skin tone, facial expressions, and hairstyles, which introduce additional semantic complexity. This complexity makes it harder for LLMs to distinguish between real and forged faces. Besides, GPT-4V demonstrates stable performance across both real and forged face datasets, achieving an accuracy of 76.7\% on real faces and 79.6\% and 76.2\% on AutoSplice and HiSD, respectively. However, DeepSeek-VL2 and Llama-3.2-Vision show a significant performance gap between real and forged face datasets, which suggests that these models lack sufficient knowledge about human faces, leading them to classify real faces as fake more frequently.

\subsection{Forgery Analysis Performance}

\begin{figure}[htbp]
  \centering
  \includegraphics[width=\linewidth]{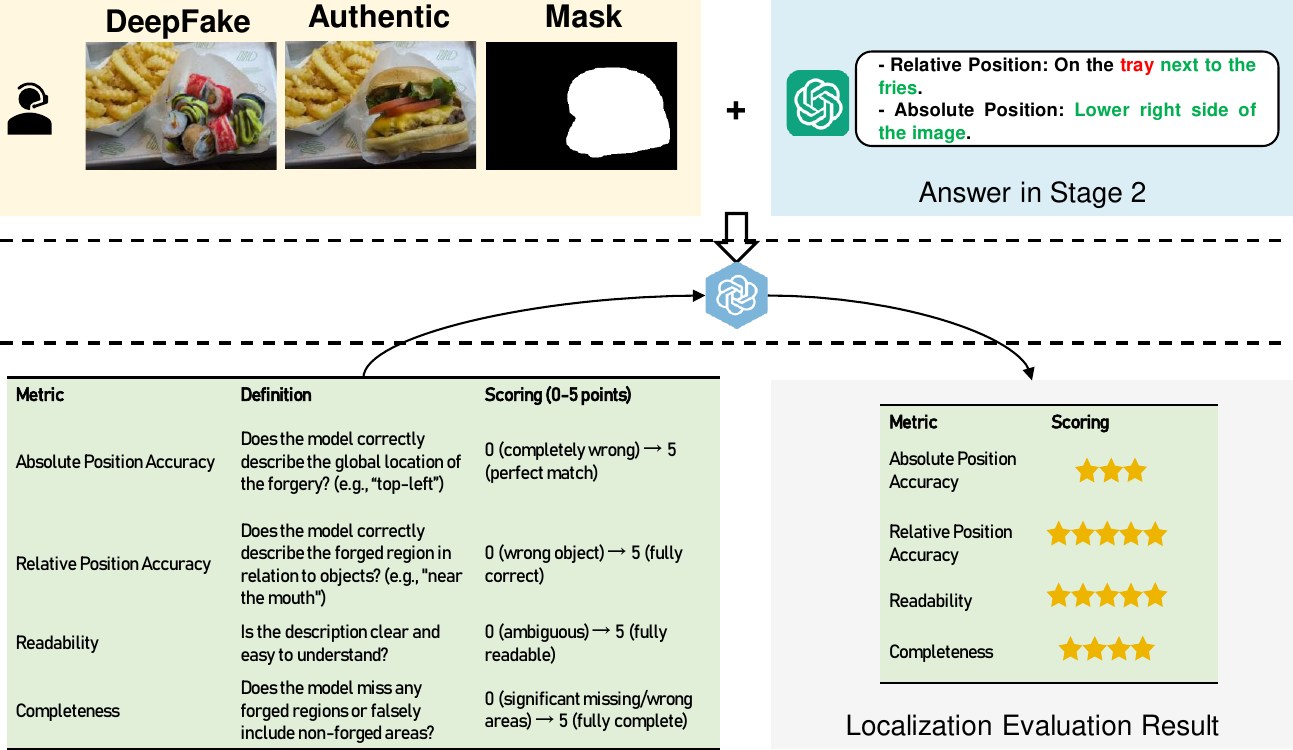}
  \caption{Examples of GPT-4o for locating forged regions. Left above: DeepFake image, authentic image, and mask image of the same picture. Right above: The answer of the LLM in Stage 2. We only need the "Location of the Tampering Area" part. Left bottom:  Metrics for localization evaluation task. We evaluate the performance of the LLM from four perspectives (absolute position accuracy, relative position accuracy, readability and completeness), each metric is rated on a 0-5 scale, where higher scores indicate better performance. Right bottom: localization evaluation result. We average the scores of four metrics as the final score.}
  \Description{Examples of GPT-4o for localization evaluation.}
  \label{fig:location_example}
\end{figure}

\begin{table}
  \caption{Comparsion of ACC (\%) in the localization task.}
  \label{tab:location auc}
\begin{tabular}{@{}lcc@{}} \toprule
 Method & Autosplice\cite{autosplice} & LaMa\cite{LaMa} \\ \midrule
  Deepseek & 30 & 28.5 \\ 
  Llama & 37.5 & 44.75 \\
  GPT4V & \textbf{66.25} & \textbf{72} \\ \bottomrule
 \end{tabular}
\end{table}

For locally forged images, we evaluate the accuracy of LLMs in localizing tampered regions. Since LLMs output natural language descriptions rather than pixel-level masks, we designed a novel evaluation framework tailored to natural language outputs. As illustrated in Figure~\ref{fig:location_example}, we employ a separate multimodal LLM to assess the localization results from Stage 2. The input to this evaluator LLM includes the textual output from the Stage 2 LLM (describing the tampered regions), the forged image, the original image, and the corresponding mask. Additionally, we provide evaluation metrics in the system prompt, as shown in the bottom-left corner of Figure~\ref{fig:location_example}. The evaluation metrics are divided into four components: Absolute Position Accuracy, Relative Position Accuracy, Readability, and Completeness. The final score is calculated as the average of these four scores, normalized to a percentage:
\begin{displaymath}
\text{Final Score} = \frac{\sum_{i=1}^{4} \text{Score}_i}{4} \times 100\%,
\end{displaymath}
where ${Score}_i$ represents the score for the i-th metric, and $Final Score$ is the overall localization accuracy. 

We evaluated the localization accuracy on two datasets: AutoSplice (based on Diffusion methods) and LaMa (based on GAN methods). GPT-4V was used as the evaluator model, and the results are summarized in Table~\ref{tab:location auc}. The finding indicates that GPT-4V achieves the highest localization accuracy, significantly outperforming DeepSeek-VL2 and Llama-3.2-Vision. This aligns with the trend observed in Stage 1, where GPT-4V's superior semantic understanding capabilities contribute to its robust performance. On the other hand, DeepSeek-VL2 shows lower accuracy on both datasets, with minimal variation between AutoSplice and LaMa. This suggests that its smaller model size limits its ability to accurately interpret semantic cues, reducing its sensitivity to differences in forgery methods. Besides, GPT-4V and Llama-3.2-Vision perform better on the LaMa dataset than on AutoSplice. This is likely because Diffusion-based methods (e.g., AutoSplice) produce smoother boundaries in tampered regions, making localization more challenging for LLMs that rely on semantic understanding. In contrast, GAN-based methods (e.g., LaMa) often introduce more noticeable artifacts, which are easier for LLMs to detect. We also notice that LLMs generally perform better in describing absolute positions than relative positions. For example, in Figure~\ref{fig:location_example}, GPT-4V correctly identifies the absolute location of a tampered region but mislabels a "hamburger wrapper" as a "tray." This indicates that while LLMs excel at high-level semantic understanding, they may struggle with fine-grained object recognition. Fine-tuning on specific datasets could mitigate this issue.

\begin{table}
  \caption{Comparsion of ACC (\%) in generation method classification. Dataset ends with (f) means this is a face dataset, otherwise a general dataset.}
  \label{tab:method acc}
\resizebox{\linewidth}{!}{
\begin{tabular}{@{}lcccccc@{}} \toprule
 \multirow{2}{*}{Method} & \multicolumn{3}{c}{Diffusion} & \multicolumn{3}{c}{GAN} \\ \cmidrule(l){2-7}
  & Autosplice\cite{autosplice} & Stable\cite{stylegan} & Autosplice(f)\cite{autosplice} & LaMa & Style & HiSD(f) \\ \midrule
 Deepseek & 6.4 & 1.7 & 5.3 & \textbf{92.4} & 93.6 & \textbf{92.3} \\
 Llama & 15.2 & 12.3 & 10.7 & 86.0 & 83.8 & 79.7 \\
 GPT4V & \textbf{94.9} & \textbf{68.9} & \textbf{70.1} & 92.2 & \textbf{95.6} & 66.5 \\ \bottomrule
 \end{tabular}
 }
\end{table}

In the generation method classification task, we evaluate the ability of multimodal LLMs to trace the forgery method used to create an image. This task is framed as a binary classification problem, where the model must choose between two generation methods: Diffusion or GAN. A correct classification is scored as 1, and an incorrect classification is scored as 0. Compared to the real vs. fake classification task, this task is more challenging because the LLM must not only understand the image but also leverage its pre-trained knowledge of Diffusion and GAN methods to establish connections between semantic features and the generation technique. Using this scoring method, we calculated the accuracy (ACC) of each LLM on datasets generated using different methods. The results are summarized in Table~\ref{tab:method acc}. Notably, GPT-4V achieves remarkable accuracy on both Diffusion and GAN datasets (94.9\% for AutoSplice and 95.6\% for StyleGAN), despite not being explicitly provided with definitions of Diffusion or GAN in the prompt. This indicates that GPT-4V effectively utilizes its pre-trained knowledge to make informed judgments about the generation method. In contrast, DeepSeek-VL2 and Llama-3.2-Vision show a strong bias toward classifying images as GAN-generated, with significantly lower accuracy on Diffusion datasets. This suggests that these models have limited pre-trained knowledge of Diffusion methods, making them more likely to default to GAN classifications. Additionally, we observe that LLMs perform worse on face datasets compared to general datasets. For example, GPT-4V's accuracy on the face version of AutoSplice drops by approximately 25\% compared to its performance on the general dataset. This further supports our earlier conclusion that face forgery detection is more challenging for LLMs due to the increased semantic complexity of human faces.

\subsection{Ablation Studies}

\begin{figure}[htbp]
  \centering
  \includegraphics[width=\linewidth]{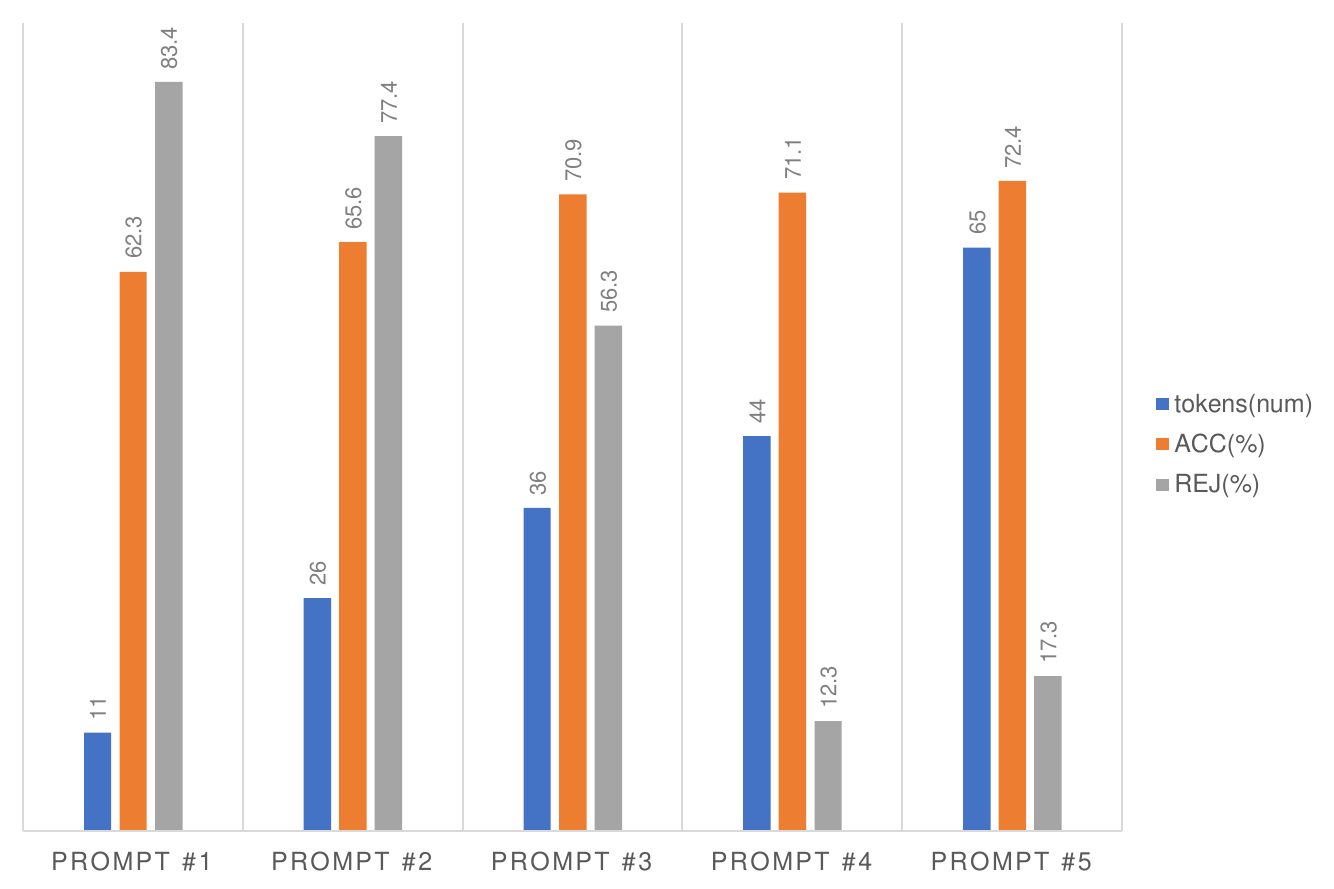}
  \caption{Comparison of different prompts for GPT4V in detecting 1,000 faces from Autosplice dataset. We sort from left to right according to the number of tokens. REJ(\%) means the rejection rate.}
  \Description{Ablation Study of Prompts.}
  \label{fig:prompt_table}
\end{figure}

\noindent \textbf{Prompt Ablation:} The quality of prompts plays a critical role in the performance of multimodal LLMs. In addition to the prompt used in our main experiments, we investigated other prompt structures and compared their effectiveness. As illustrated in Figure~\ref{fig:prompt_ablation}, inspired by LangGPT and OpenAI's official documentation, our prompt design is based on five principles: Profile, Goal, Constraint, Workflow, and Style. From top to bottom, the prompts not only increase in token count but also incorporate more of these principles. To evaluate the impact of these prompts on forgery detection tasks, we quantitatively compared their performance on 1,000 face images from the AutoSplice dataset. Figure~\ref{fig:prompt_table} reports the accuracy (ACC) and the rejection rate(REJ) for GPT-4V across all five prompts. The results reveal a positive correlation between token count and accuracy, indicating that detailed task descriptions and additional contextual information help bring forth the power of semantic knowledge of LLMs in forgery detection tasks. We also find that prompts that directly request image forgery detection, such as Prompt \#1 and Prompt \#2, exhibit higher rejection rates (83.4\% for Prompt \#1 and 77.4\% for Prompt \#2). In contrast, Prompt \#4 and Prompt \#5, which incorporate more principles like Profile and Goal, show significantly lower rejection rates. This suggests that providing a clear profile of the task and defining specific goals are crucial for reducing refusal rates and improving the performance of LLMs. Additionally, it should be noted that while longer prompts improve accuracy, they also increase the computational cost of running LLMs. We ultimately selected Prompt \#4 for Stage 1 in our experiments to balance performance and cost. For Stage 2, we followed similar design principles but added more detailed descriptions and clues about the tasks. And we finally utilized the prompt shown in Figure~\ref{fig:prompt}.

\begin{figure}[htbp]
  \centering
  \includegraphics[width=0.75\linewidth]{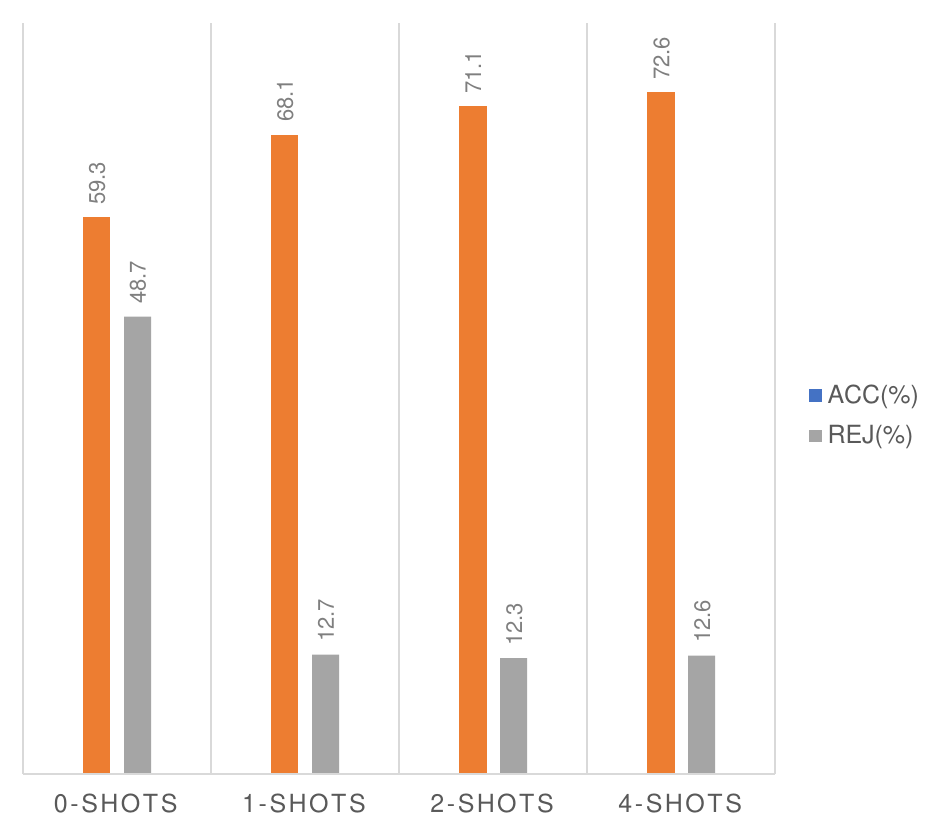}
  \caption{Exemplar sensitivity analysis for GPT4V in detecting 1,000 faces from Autosplice dataset. Specifically, we wrote 10 total exemplars. For k-shot learning, we randomly sample k=(0, 1, 2, 4) out of 10 exemplars. }
  \Description{Ablation Study of Prompts.}
  \label{fig:examplar}
\end{figure}

\noindent \textbf{Sensitivity to Exemplars:} Incorporating exemplars into prompts can significantly enhance the ICL capabilities of LLMs. To better understand the sensitivity of LLMs to different exemplars, we conduct a sensitivity analysis. Specifically, we design a total of 10 exemplars and perform k-shot learning experiments by randomly sampling k = (0, 1, 2, 4) exemplars from the pool three times each. The average performance across these trials is then evaluated on a randomly selected set of 1,000 face images from the AutoSplice dataset. We use prompt \#4 in Figure~\ref{fig:prompt_ablation}. The results of our sensitivity analysis are summarized in Figure~\ref{fig:examplar}. First, we observed that increasing the number of shots (exemplars) has a more pronounced effect on reducing rejection rates (REJ) than on improving accuracy (ACC). Without exemplars, GPT-4V tends to refuse to answer questions related to face and forgery detection. By adding exemplars, LLMs can better generalize to forgery detection tasks, as the exemplars provide contextual guidance and reduce ambiguity. However, we also found that the marginal improvement in ACC diminishes as the number of exemplars increases. For example, while moving from 0-shot to 1-shot learning yields a significant boost in performance (8.8\% improvement), the gains from 2-shot to 4-shot learning are less pronounced(1.5\% improvement). Additionally, using more exemplars increases computational costs, which is an important consideration for large-scale evaluation. Based on these findings, we ultimately decided to use 2-shot learning for both Stage 1 and Stage 2 of our experiments. This trade-off between performance improvement and computational efficiency ensures that the LLMs benefit from contextual guidance without incurring excessive costs.

\subsection{Improvements}

\begin{figure}[htbp]
  \centering
  \includegraphics[width=\linewidth]{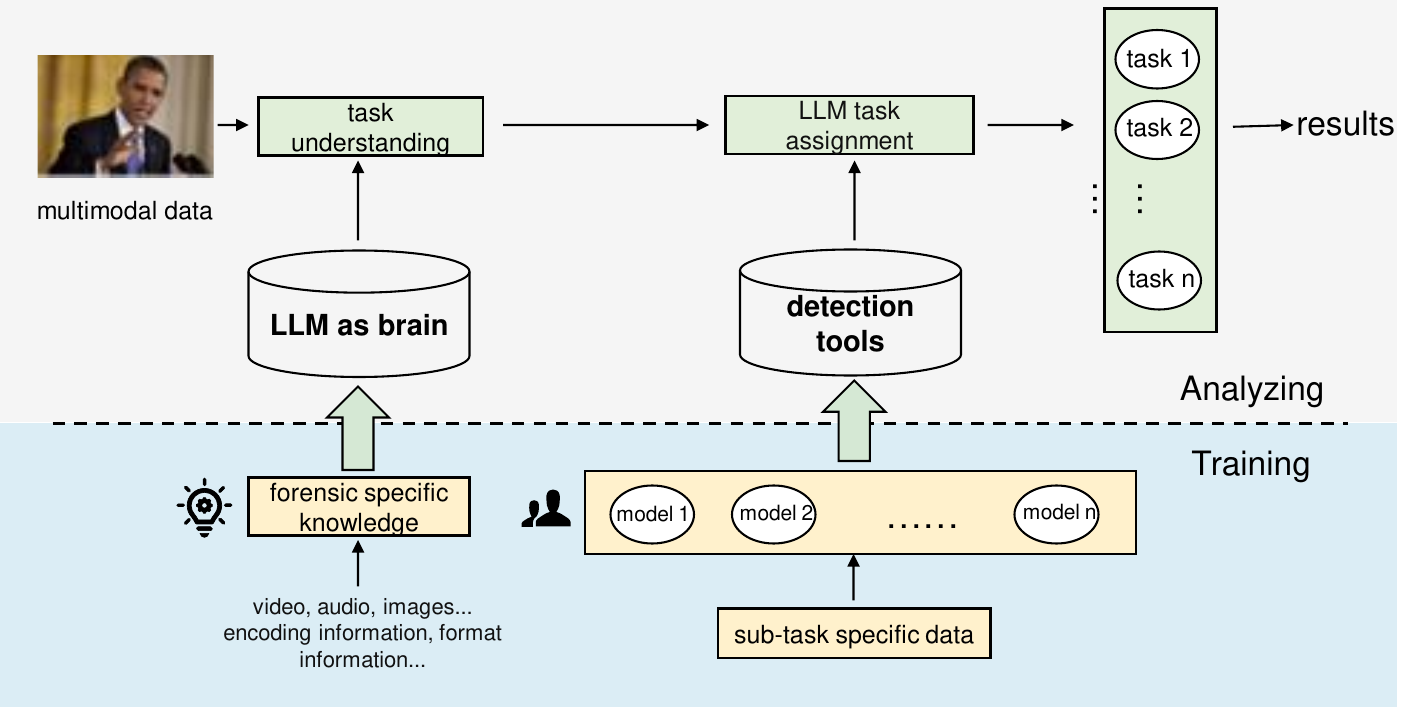}
  \caption{Potential improvement for future forensic detection. LLM can act as a connector between multimodal data and downstream forensic detecting tasks, which can assign different tools and models for different sub-tasks, achieving fine-grained forgery analysis.}
  \label{fig:llm_agent}
\end{figure}

So far, our experiments have focused on evaluating the performance of multimodal LLMs on image-based Deepfake detection tasks. However, with the rapid advancement of generative AI technologies, AI-generated content in other modalities, such as video and audio, has also seen significant progress. Detecting Deepfakes in videos and audio presents unique challenges, such as temporal consistency in videos and spectral patterns in audio. While LLMs have demonstrated strong semantic understanding capabilities in image analysis, their application to video and audio forgery detection remains largely unexplored. Future work could investigate how to effectively integrate LLMs with multimodal data pipelines, leveraging their ability to interpret complex semantic relationships across different modalities. 

Another potential improvement lies in the combination of the strengths of large and small models. For example, we can create a hybrid system that leverages the generalization capabilities of LLMs and the precision of specialized small models or tools to achieve fine-grained forgery analysis. Figure~\ref{fig:llm_agent} illustrates an exploratory framework. In this setup, data is preprocessed and fed into an LLM, which acts as a connector and task allocator. Based on its pre-trained knowledge and semantic understanding, the LLM assigns specific tasks to downstream small models or tools. This approach capitalizes on the strengths of both large and small models: the LLM provides broad semantic understanding and task coordination, while the small models offer high accuracy and efficiency in specialized tasks. We hope that such a framework could significantly enhance the robustness and scalability of DeepFake detection systems.

\section{CONCLUSION}

In this work, we investigate the potential of multimodal LLMs for AIGC detection and forensic analysis. We explore the application of a two-stage framework to facilitate a comprehensive and systematic analysis of potentially forged images. The findings reveal that LLMs, particularly GPT-4V, exhibit a significant potential for analyzing AIGC both qualitatively and quantitatively. Moreover, LLMs demonstrate remarkable versatility across multiple datasets without necessitating explicit training for specific DeepFake contexts, primarily due to their ability to draw on a rich repository of semantic knowledge. The nuanced design of the prompts and the strategic incorporation of sourced examples appear crucial in optimizing model performance while mitigating refusal rates. Our work makes LLMs a versatile and practical tool for diverse real-world applications. For future work, strategies for expanding LLM applications to cover other media formats such as video and audio remain an exciting avenue of research. Through these efforts, LLMs could significantly improve the efficacy of forensic detection of contemporary media against the widespread threat of DeepFakes.

\bibliographystyle{ACM-Reference-Format}
\bibliography{canGPT}

\end{document}